\documentclass{article}

\usepackage[main, final]{iaseai26}

\usepackage[utf8]{inputenc} 
\usepackage[T1]{fontenc}    
\usepackage{hyperref}       
\usepackage{url}            
\usepackage{booktabs}       
\usepackage{amsfonts}       
\usepackage{nicefrac}       
\usepackage{microtype}      
\usepackage{longtable}
\usepackage{multirow}
\usepackage{tabularx}
\usepackage{array,makecell}
\usepackage[table]{xcolor}
\usepackage{graphicx}
\usepackage{soul}

\newcolumntype{Y}{>{\raggedright\arraybackslash}X}

\title{With Great Capabilities Come Great Responsibilities: Introducing the Agentic Risk \& Capability Framework for Governing Agentic AI Systems}

\author{
  Shaun Khoo, Jessica Foo\thanks{Equal contribution} \\
  GovTech Singapore\\
  \And
  Roy Ka-Wei Lee\\
  Singapore University of Technology and Design
}

\begin{document}
\maketitle
\typeout{}

\begin{abstract}
Agentic AI systems present both significant opportunities and novel risks due to their capacity for autonomous action, encompassing tasks such as code execution, internet interaction, and file modification. This poses considerable challenges for effective organizational governance, particularly in comprehensively identifying, assessing, and mitigating diverse and evolving risks. To tackle this, we introduce the Agentic Risk \& Capability (ARC) Framework, a technical governance framework designed to help organizations identify, assess, and mitigate risks arising from agentic AI systems. The framework's core contributions are: (1) it develops a novel capability-centric perspective to analyze a wide range of agentic AI systems; (2) it distills three primary sources of risk intrinsic to agentic AI systems - components, design, and capabilities; (3) it establishes a clear nexus between each risk source, specific materialized risks, and corresponding technical controls; and (4) it provides a structured and practical approach to help organizations implement the framework. This framework provides a robust and adaptable methodology for organizations to navigate the complexities of agentic AI, enabling rapid and effective innovation while ensuring the safe, secure, and responsible deployment of agentic AI systems. Our framework is open-sourced \href{https://govtech-responsibleai.github.io/agentic-risk-capability-framework/}{here}.
\end{abstract}

\section{Introduction}

OpenAI dubbed 2025 the "year of the AI agent” \citep{hamilton2025_agents}, a prediction that quickly proved prescient. Major AI companies launched increasingly powerful systems that allowed large language model ("LLM") agents to reason, plan, and autonomously execute tasks such as code development or web surfing. However, this surge in agent-driven AI innovation also brought renewed scrutiny to these systems' safety and security risks. Recent research \citep{chiang2025_harmful, kumar2025_aligned, yu2025_safety} demonstrated that LLM agents are more prone to unsafe behaviors than their base models. Moreover, governing agentic systems presents unique challenges compared to traditional LLM systems - they have the autonomy to execute a wide variety of actions, thereby introducing a significantly broader range of risks. This makes comprehensive identification, assessment, and mitigation more challenging, thus hindering effective organizational governance. Although conducting in-depth and customized risk assessments for each agentic system is possible as an interim measure, it is unsustainable in the long run.

The Agentic Risk \& Capability ("ARC") framework aims to tackle this problem as \textbf{a technical governance framework for identifying, assessing, and mitigating the safety and security risks of agentic systems}. It examines where and how risks may emerge, contextualizes the agentic system's risks given its domain, use case, and organizational context, and recommends practical and technical controls for mitigating these risks. While the ARC framework is not a panacea to the complex challenges of governing agentic systems, it offers a strong foundation upon which organizations can manage the plethora of risks in a systematic, scalable, and adaptable manner.

\section{Existing Literature on Agentic AI Governance}

Although regulatory frameworks such as the EU AI Act \citep{eu_ai_act} and the NIST Risk Management Framework \citep{nist_ai_rmf} articulate clear overarching principles and guidelines for managing AI risks, they do not examine specific technical measures for identifying, assessing, and managing risks. Our paper aims to contribute to the \textbf{technical AI governance} field by developing "technical analysis and tools for supporting the effective governance of AI" \citep{reuel2025openproblemstechnicalai}. For agentic AI, \citet{raza2025trismagenticaireview} adapted the AI Trust, Risk, and Security Management (TRiSM) framework to LLM-based multi-agent systems. It provides generalized metrics and controls across a spectrum of risks, but does not tackle the practical problems of contextualizing risks for a given agentic system to be deployed. Another approach, proposed by \citet{engin2025adaptivecategoriesdimensionalgovernance}, is dimensional governance through tracking AI systems along three dynamic axes (decision authority, process autonomy, and accountability), introducing controls when systems shift across critical thresholds. While conceptually appealing, its effectiveness relies on accurately quantifying the dimensions and calibrating the thresholds, both of which are hard to operationalize. More cybersecurity-oriented frameworks include the MAESTRO framework \citep{huang2025_resilience}, OWASP's white paper on agentic AI risks \citep{owasp2025_agentic_threats}, and NVIDIA's taint tracing approach \citep{nvidia2025_autonomy} which utilize threat modelling to uncover security threats (e.g. data poisoning, agent impersonation). However, this is highly complex, especially for developers untrained in cybersecurity, and the controls rely heavily on human oversight. 

Benchmarks help to assess how risky agentic systems are, and there are several safety and security benchmarks which outline test scenarios or tasks that reveal specific risk behaviors of the agentic system. For example, Agent Security Bench \citep{zhang2025agentsecuritybenchasb}, CVEBench \citep{zhu2025cvebenchbenchmarkaiagents}, RedCode \citep{guo2024_redcode}, AgentHarm \citep{andriushchenko2025agentharmbenchmarkmeasuringharmfulness}, AgentDojo \citep{debenedetti2024agentdojodynamicenvironmentevaluate} assess whether LLMs can complete multi-step cybersecurity attacks or harmful tasks like fraud, but they do not help developers identify the full range of risks and attack scenarios of their specific applications. Tool-based benchmarks, such as APIBench \citep{patil2023gorillalargelanguagemodel}, ToolSword \citep{ye2024toolswordunveilingsafetyissues}, and ToolEmu \citep{ruan2024identifyingriskslmagents} measure the performance and safety of LLMs in utilizing tools like \texttt{bash}, but omit risks unrelated to tool use (e.g. misaligned LLMs).

For mitigating risks, AI control has emerged as a paradigm in preventing misaligned AI systems from causing harm \citep{greenblatt2024aicontrolimprovingsafety}. Rather than relying solely on training techniques to shape model behavior, AI control focuses on designing mechanisms like monitoring and human oversight to constrain AI systems. For instance, Progent \citep{shi2025progentprogrammableprivilegecontrol} and AgentSpec \citep{wang2025agentspeccustomizableruntimeenforcement} introduce a language for flexibly expressing privilege control policies that are applied at runtime. The UK AI Security Institute advocates for AI control levels, derived from evaluating frontier LLMs' threat model-specific capabilities \citep{korbak2025evaluatecontrolmeasuresllm}. OpenAI shared best practices like constraining the agent's action spaces and ensuring attributability \citep{openai_practices_governing_agentic}, while Google emphasized a hybrid defense-in-depth strategy that combines deterministic security measures with dynamic, reasoning-based defenses \citep{google_approach_secure_ai_agents}. Similarly, \citet{beurerkellner2025designpatternssecuringllm} propose six design patterns for building AI agents with provable resistance to prompt injections. However, these works are either too narrow (i.e., specific to application) or too broad (i.e., high-level, conceptual) to effectively operationalize within organizations. 


\section{Capabilities of an Agentic System}

Effective governance requires distinguishing between safer and riskier systems and implementing a differentiated approach to manage them. Applying this to agentic AI governance, beyond analyzing the components of an agent (i.e. the LLM, instructions, tools, and memory) and the design of the agentic system (i.e. agentic architecture, access controls, and monitoring), \textbf{the ARC framework adopts the novel approach of also analyzing agentic AI systems by their capabilities.}

\textbf{By capabilities, we refer to the actions that the agentic system can autonomously execute over the tools and resources it has access to, whether it be running code, searching the internet, or modifying documents.} This is the complement of affordances (as defined by \citet{gaver_affordances}), which are properties of the external environment that enable actions. In our view, the components and design of agentic systems (see Sections \ref{sec:components} and \ref{sec:design}) are \emph{affordances}, while executing code or altering agent permissions are examples of \emph{capabilities}, which we cover in Section \ref{sec:capabilities}. Addressing both aspects is essential for the effective governance of agentic systems.

There are three key advantages of adopting a capability lens in agentic AI governance. 

\begin{enumerate}

    \item \textbf{Capabilities offer a more holistic unit of analysis than analyzing specific tools}. There are numerous tools that facilitate similar actions (e.g. Google SERP, Serper, SerpAPI, Perplexity Search API), and conversely, a single tool can enable a wide array of actions (e.g. GitHub’s Model Context Protocol ("MCP") server enabling code commits, reading of pull requests etc.) - a point also made by \citet{gaver_affordances} on affordances. Given the sheer diversity and rapid development of MCPs, prescribing specific controls for each and every tool used would be too granular, and lead to obsolete, inconsistent, and overly restrictive controls. 

    \item \textbf{Adopting a capability lens allows for differentiated treatment in a scalable manner}. Systems with more capabilities are inherently riskier and necessitate more stringent controls, particularly when these capabilities have a significant impact on the system. By deconstructing a system into its constituent capabilities, we can ensure that riskier systems receive greater scrutiny while enabling low-risk systems to proceed with a lighter touch.

    \item \textbf{Risks arising from actions is intuitive to laypersons, which is vital for effective contexualization}. Technical approaches often run the risk of being esoteric, which hampers adoption and limits flexibility. By being more accessible to the average person, the capability lens enables organizations to be more flexible in adapting to new developments and risks.
    
\end{enumerate}


\section{Agentic Risk \& Capability Framework}

In this section, we explain each part of the ARC framework - the elements, risks, and controls - in detail. We also provide a visual summary of the entire framework in Figure 1.

\begin{figure}[h]
    \centering
    \includegraphics[width=1\linewidth]{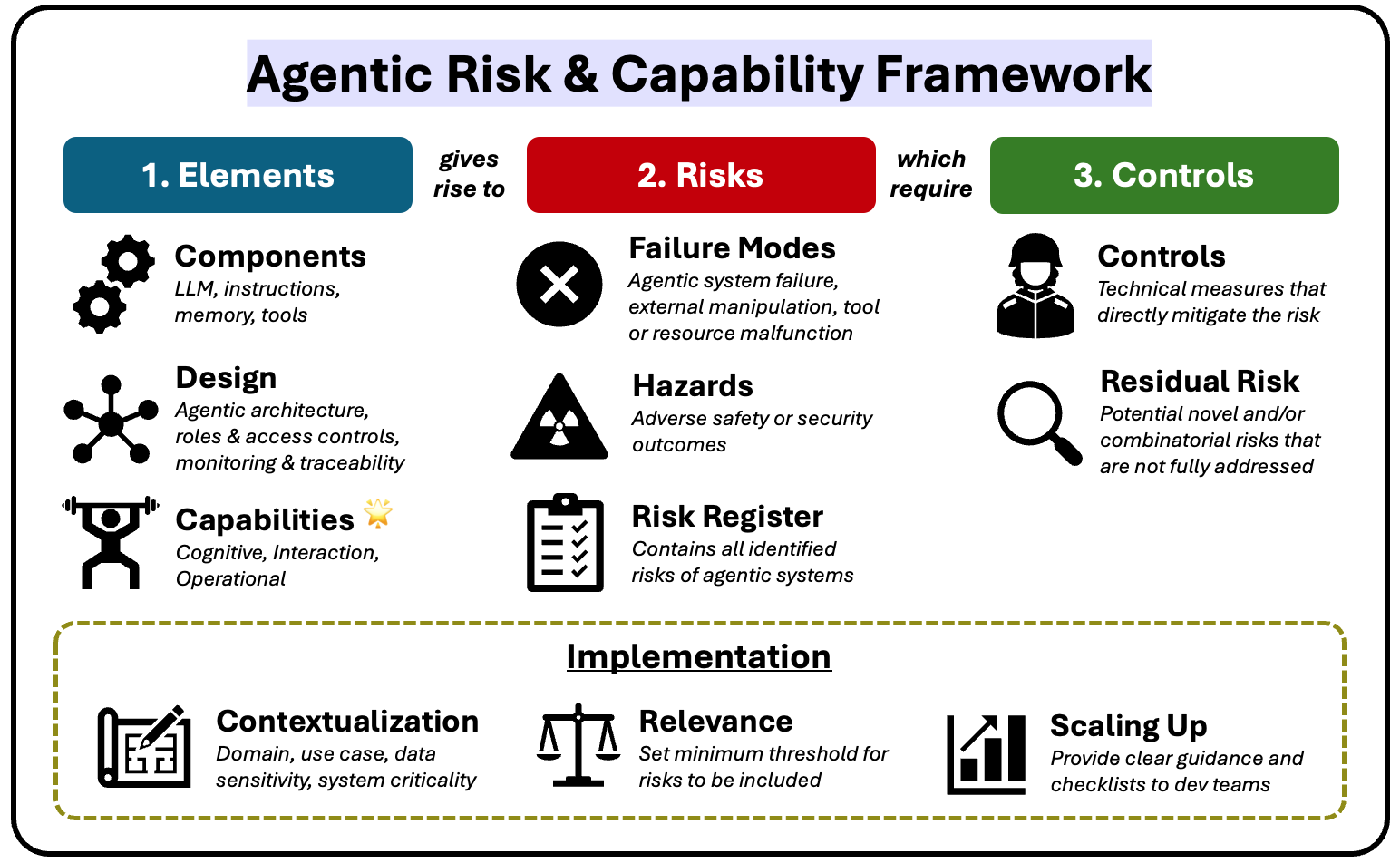}
    \caption{Overview of the ARC Framework}
    \label{fig:arc_framework}
\end{figure}

\subsection{Elements of Agentic Systems}
\label{sec:elements}

Across all agentic systems, there are three indispensable elements to examine: components of an agent, design of the agentic system, and the capabilities of the agentic system.

\subsubsection{Components}
\label{sec:components}

Components are essential parts of a single, standalone agent. Here, we synthesize prevailing agreement on the key components of an agent from various sources, such as OpenAI \citep{openai2025_agents}.

\begin{itemize}

    \item \textbf{LLM}: The LLM is the central reasoning engine that processes instructions, interprets user inputs, and generates contextually appropriate responses by leveraging its trained language understanding and generation capabilities.

    \item \textbf{Tools}: Tools enable LLMs to interact with the external environment, be it editing files, querying databases, controlling devices, or accessing APIs. This is facilitated by MCP servers, which provide LLMs a consistent interface to discover and utilize a variety of tools.

    \item \textbf{Instructions}: Instructions are the blueprint which defines an agent's role, capabilities, and behavioral constraints, ensuring it operates within intended parameters and maintains its performance across different scenarios. 

    \item \textbf{Memory}: The memory or knowledge base component provides the agent with contextual awareness and information persistence, enabling it to maintain coherent conversations, learn from past interactions, and access relevant facts without requiring constant re-instruction.

\end{itemize}

\subsubsection{Design}
\label{sec:design}

We now broaden our perspective to examine how agentic AI systems are assembled from individual agents from a system design perspective.

\begin{itemize}

    \item \textbf{Agentic Architecture}: The agentic architecture defines how multiple agents are interconnected, coordinated, and orchestrated to collectively solve complex tasks that exceed individual agent capabilities, including patterns like hierarchical delegation, parallel processing, or sequential handoffs between specialized agents. Different architectures result in varying levels of system-wide risk, and these need to be considered carefully. Similarly, the protocols \citep{a2a} by which agents communicate may also give rise to security risks. 

    \item \textbf{Roles and Access Controls}: Roles and access controls establish differentiated responsibilities and permissions across agents within the system, ensuring that each agent operates within appropriate boundaries while being able to fulfill its designated function. This is critical because it limits unauthorized actions, contains the blast radius of potential failures or security breaches, and enables the system to maintain reliability even when individual agents may be compromised or behave unexpectedly.

    \item \textbf{Monitoring and Traceability}: Monitoring and traceability enable visibility into agentic system behavior, interactions, and decision-making pathways, allowing developers and operators to understand what agents are doing, why they made particular choices, and how outcomes were produced. This is essential for post-hoc debugging, real-time anomaly detection, and establishing accountability particularly when agents operate with a degree of autonomy or interact with sensitive systems and data.

\end{itemize}

\subsubsection{Capabilities}
\label{sec:capabilities}

We see three broad categories of capabilities - cognitive, interaction, and operational - and break it down into more granular capabilities.

\ul{\textbf{Cognitive capabilities}} encompass the agentic AI system's internal "thinking" skills – how it analyses information, forms plans, learns from experience, and monitors its own performance.

\begin{itemize}
    
    \item \textbf{Planning \& Goal Management}: The capability to develop detailed, step-by-step, and executable plans with specific tasks in response to broad instructions. This includes prioritizing activities based on importance and dependencies between tasks, monitoring how well its plan is working, and adjusting when circumstances change or obstacles arise.
    
    \item \textbf{Agent Delegation}: The capability to assign subtasks to other agents and coordinate their activities to achieve broader goals. This includes identifying which components are best suited for specific tasks, issuing clear instructions, managing inter-agent dependencies, and monitoring performance or failures.
    
    \item \textbf{Tool Use}: The capability to evaluate available options and choose the best tool for specific subtasks. This requires agents to understand the capabilities and limitations of different tools and match them appropriately to the tasks.
    
\end{itemize}

\ul{\textbf{Interaction capabilities}} describe how the agentic AI system exchanges information with users, other agents, and external systems. These capabilities below are broadly differentiated based on how and what they interact with.

\begin{itemize}
    \item \textbf{Natural Language Communication}: The capability to fluently and meaningfully converse with human users, handling a wide range of situations such as explaining complex topics, generating documents or prose, or discussing issues with human users.

    \item \textbf{Multimodal Understanding \& Generation}: The capability to take in image, audio, or video inputs and / or generate image, audio, or video outputs. This includes analyzing visual information, transcribing speech, or creating multimedia content as needed.

    \item \textbf{Official Communication}: The capability to compose and directly publish communications that formally represent an organization to external parties (e.g. customers, partners, regulators, courts, media) via approved channels and formats without human oversight. 

    \item \textbf{Business Transactions}: The capability to execute transactions that involve exchanging money, services, or commitments with external parties. It can process payments, make reservations, and handle other business transactions within authorized limits. 

    \item \textbf{Internet \& Search Access}: The capability to access and search the Internet for knowledge resources, especially for up-to-date information to provide more accurate answers. 

    \item \textbf{Computer Use}: The capability to directly control a computer interface by moving the mouse, clicking buttons, and typing on behalf of the user. It can navigate applications and perform tasks that require interacting with graphical user interfaces.

    \item \textbf{Other Programmatic Interfaces}: The capability to interact with external systems through APIs, SDKs, or backend services. This includes sending and receiving data via RESTful APIs, pushing code to a remote repository, or invoking cloud services to retrieve or manipulate information from other systems. 
    
\end{itemize}

\ul{\textbf{Operational capabilities}} focus on the agentic AI system's ability to execute actions safely and efficiently within its operating environment.

\begin{itemize}

    \item \textbf{Code Execution}: The capability to write, execute, and debug code in various programming languages to automate tasks or solve computational problems. 

    \item \textbf{File \& Data Management}: The capability to create, read, modify, organize, convert, query, and update information across both unstructured files (e.g. PDFs, Word docs, spreadsheets) and structured data stores (e.g. SQL/NoSQL databases, data warehouses, vector stores). 

    \item \textbf{System Management}: The capability to adjust system configurations, manage computing resources, and handle technical infrastructure tasks. This includes monitoring system performance, securely handle authentication information and access controls, and making optimizations as needed while maintaining security best practices. 
    
\end{itemize}

\subsection{Part 2: Risks of Agentic Systems}

The next part involves detailing how the risks materialize from the elements of an agentic system as described in \ref{sec:elements}. This comprises two key aspects: the failure mode, which outlines how the system fails, and the hazard, which describes the resulting impact.

\subsubsection{Failure Modes}

First, we specify three general modalities in which agentic systems may fail:

\begin{itemize}
    \item \textbf{Agent Failure}: The agent itself fails to operate as intended due to poor performance, misalignment, or unreliability.
    \item \textbf{External Manipulation}: Malicious actors cause or trick the agent to deviate from its intended behavior.
    \item \textbf{Tool or Resource Malfunction}: The tools or resources utilized by the agentic system fail, are compromised, or are inadequate.
\end{itemize}

\subsubsection{Hazards}

Second, we list a range of safety and security hazards which may result from these failures. Note that this distinction serves solely as a heuristic for comprehensive risk identification and should not be interpreted as a rigid taxonomic principle.

\begin{table}[h]
\centering
\caption{Hazard Categories by Type}
\setlength{\tabcolsep}{6pt}
\renewcommand{\arraystretch}{1.2}

\rowcolors{3}{gray!6}{white}

\begin{tabularx}{\linewidth}{
  >{\centering\arraybackslash}m{2cm}   
  >{\raggedright\arraybackslash}p{3cm} 
  >{\raggedright\arraybackslash}X        
}
\toprule
\textbf{Type} & \textbf{Hazard Category} & \textbf{Description} \\
\midrule

\cellcolor{white} & Data (files, databases) &
Failures can lead to data breaches, integrity attacks, PII exposure, or ransomware, where sensitive information is exfiltrated, corrupted, or held hostage. \\

\cellcolor{white} & Application &
This category includes system failures, service disruptions, unintended use of applications, backdoor access, or resource exploitation, compromising the functionality and security of the software. \\

\cellcolor{white}{\multirow{-4}{*}{\textbf{Security}}} & Infrastructure \& network &
Denial of service (DoS/DDoS) attacks, man-in-the-middle (MitM) attacks, network eavesdropping, or lateral access, all of which can disrupt or compromise the underlying network and infrastructure. \\

\cellcolor{white} & Identity \& access management &
Unauthorized control, impersonation of credible roles, or privilege escalation, allowing attackers to gain elevated access or control over systems. \\

\midrule

\cellcolor{white}  & Illegal and CBRNE activities &
This includes agents facilitating or engaging in CBRNE-related activities or other types of criminal offenses, such as fraud, scams, or smuggling. \\

\cellcolor{white}  &
Discriminatory or hateful content &
This category is aimed at unsafe and discriminatory content, especially incendiary hate speech and slurs, as well as biased decisions. \\

\cellcolor{white} & Inappropriate content &
This refers to the generation of content that is vulgar, violent, sexual, promotes self-harm, or encourages illegal activities, leading to reputational harm and erosion of trust in the system. \\

\cellcolor{white}{\multirow{-5}{*}{\makecell[c]{\textbf{Safety}}}} & Compromise user safety &
Failures can directly endanger users, for example, through the propagation of inaccurate information or the execution of actions that lead to physical or psychological harm. \\

\cellcolor{white} & Misrepresentation &
This outcome involves the propagation and dissemination of wrong and inaccurate information, or cascading failures where inaccurate information is not corrected, leading to further errors and a loss of trust. \\

\bottomrule
\end{tabularx}
\end{table}

\subsubsection{The Risk Register}

The Risk Register consolidates all the risks identified through the ARC framework, and \textbf{serves as the organization's reference list of safety and security risks of agentic systems}. By design, each risk in the Risk Register should (1) originate from an element (components, design, or capabilities), (2) satisfy a failure mode (agent failure, external manipulation, tool or resource malfunction), and (3) result in at least one of the safety or security hazards listed in the table above. We generally recommend phrasing risks in a consistent manner to aid validation and understanding.

To demonstrate how this works in practice, we provide three examples below:

\fbox{%
  \parbox{0.98\linewidth}{%
    \begin{flushleft}

    \textbf{Example 1}: “Overwhelming the database with poor, inefficient, or repeated queries” is a \ul{security risk (application, infrastructure)} caused by \ul{agent failure} of the \ul{File \& Data Management capability}. 
    \newline
    \newline
    \textbf{Example 2}: “Opening vulnerabilities to prompt injection attacks via malicious websites” is a \ul{security and safety risk (all)} caused by \ul{external manipulation} of the \ul{Internet \& Search Access capability}.
    \newline
    \newline
    \textbf{Example 3}: “Poorly implemented tools may not correctly verify user identity or permissions when executing privileged actions” is a \ul{security risk (identity \& access management)} caused by \ul{tool or resource malfunction} of the \ul{tools component} in an agent.

    \end{flushleft}
  }%
}

Although combining the element, failure mode, and hazard can help in brainstorming potential risks to agentic systems, not all of them will be correct. For instance, tool or resource malfunction for the instructions component is not really a sensible risk. As such, organizations should exercise discretion in deciding what risks to be included in the Risk Register - one helpful criteria is to keep only risks which are supported by academic research or industry case studies.

For illustrative purposes, we provide a draft Risk Register in Appendix \ref{app:risk-register-risks}, covering most of the major risks associated with agentic systems, and with each risk backed by a real example or academic study. This can serve as a useful starting point for organizations, though it will need continual updating as the space of agentic AI develops and matures. 

\subsection{Part 3: Controls for Agentic Systems}

The last part provides guidance on how these risks can be mitigated through technical controls. However, given the rapidly evolving field of agentic AI, there is likely to be significant residual risk even after several controls have been implemented. We discuss both below.

\subsubsection{Technical Controls}

Within the Risk Repository, \textbf{each risk comes with a set of recommended technical controls} which aim to either (i) reduce the potential impact by limiting the scope or severity of a failure, or (ii) decrease the likelihood of a specific failure mode occurring. This makes the logical connection between risks and controls clear and intuitive. Controls are categorised into three levels based on criticality: Cardinal controls (Level 0) are fundamental requirements that must be adopted as is; Standard controls (Level 1) should be adopted or adapted meaningfully; and Best Practice controls (Level 2) are recommended for high-risk systems. This tiered approach enables organisations to prioritise control implementation based on their risk tolerance and resource constraints.

We provide an example of the technical controls for a specific risk below:

\fbox{%
  \parbox{0.98\linewidth}{%
    \begin{flushleft}

    \textbf{Risk}: “Opening vulnerabilities to prompt injection attacks via malicious websites” is a \ul{security and safety risk (all)} caused by \ul{external manipulation} of the \ul{Internet \& Search Access capability}.
    \newline
    \newline
    \textbf{Control 1}: Implement input guardrails to detect prompt injection or adversarial attacks
    \newline
    \textbf{Control 2}: Implement escape filtering before including web content into prompts
    \newline
    \textbf{Control 3}: Use structured retrieval APIs for searching the web rather than through web scraping
    
    \end{flushleft}
  }%
}

It is important to note that not all controls are unique; some may overlap due to targeting similar failure modes or aiming to limit the "blast radius" of a particular security or safety outcome. This is especially true of capabilities which create new vectors for prompt injection attacks.

In our draft Risk Register in the appendix, we also provide a tentative list of recommended controls for each risk to help organizations get started.

\subsubsection{Residual risks}

Agentic AI and LLMs is a rapidly developing space, and it is unlikely that any list of technical controls can credibly claim to entirely neutralize all potential threats. This makes it crucial to evaluate the residual risk - the remaining risk after controls have been applied - to uncover gaps and to assess the overall level of risk in the agentic system. If the residual risk is deemed unacceptable, further measures, both technical and otherwise, must be implemented to reduce it to an acceptable level. 

Identifying residual risks is intrinsically difficult as it is very dependent on the specifics of the agentic system, but common ones include inherent weaknesses of the technical controls (for example, prompt injection guardrails that are trained on past jailbreaks may not generalize well to detect novel attacks) or combinatorial risks which arise from the interaction of two or more capabilities.

\subsection{Implementation}

A well-known adage is ``Policy is implementation and implementation is policy'' \citet{ho2010_implementation}, and this is resoundingly true for AI governance. The ARC framework is designed to be implemented by organizations, and specifically by centralized governance teams that are responsible for managing the risks of AI and agentic systems. This subsection highlights three key steps that governance teams need to take when implementing the ARC framework in their organization.

\subsubsection{Contextualizing Risks}

Contextualizing risk involves two primary dimensions: determining the degree of impact and the degree of likelihood. We recommend a five-point scale for both, with impact ranging from minimal to catastrophic and likelihood ranging from very likely to very rare. This assessment must be carefully contextualized as the implications for a small enterprise in the manufacturing sector differ greatly from those of a multinational corporation in the finance industry or even a governmental entity. 

The degree of impact will vary significantly depending on how and where the agentic system is used. For instance, a hallucination in marketing copy might be tolerable, but in a legal context, it would be entirely unacceptable. Some criteria to consider when estimating the impact include:

\begin{itemize}
    \item \textbf{Domain}: The sensitivity and criticality of the domain are paramount. For instance, risks in medical or educational domains are more critical than those in less sensitive areas.
    \item \textbf{Use Case}: What the agentic system is used for. While office productivity tools might present straightforward risks, systems involved in hiring or performance assessments carry more sensitive and potentially impactful consequences.
    \item \textbf{Data Sensitivity}: The level of sensitivity or confidentiality of the data being processed. Systems handling highly sensitive data naturally pose greater risks if compromised.
    \item \textbf{System Criticality}: For governmental or critical infrastructure systems, the impact of a system failure can be severe and widespread, necessitating a higher level of scrutiny.
\end{itemize}

Assessing the degree of likelihood will depend a lot on the identified failure mode and the probability of that failure mode occurring, considering factors like the ease of replication or the level of access required for a successful attack. Although this is slightly less context-dependent, there are some factors like the organization’s general security (physical and cyber) measures that may limit how exploitable an agentic system can be.

\subsubsection{Establish Relevance Threshold}

Organizations must then establish a minimum threshold for both impact and likelihood to determine which risks are relevant to the specific agentic system. Any risks that remain above this relevance threshold will then require explicit mitigation through the controls described in Part 3 of the framework. This threshold is contingent upon the organization's overall risk appetite - some enterprises may set a higher threshold to keep the number of relevant risks small, while more conservative organizations might choose a lower threshold to require more risks to be directly managed. 

\subsubsection{Scaling Up}

To streamline implementation, organizations should provide simple forms or checklists for developers to declare system capabilities, relevant risks, and technical controls, which can then be validated and audited by a central governance team. This standardization also helps in providing an organization-wide view of risk exposures and control adoption. 

Another critical aspect is continual updating of the Risk Register, especially as new threats or regulatory changes emerge. Organizations need to define a regular cadence for reviewing the risks and controls in the Risk Register, and updating them to keep up with the latest developments.

\section{Worked Examples}

In this section, we apply the ARC framework to two stylized agentic systems to demonstrate how the framework would help in practice to identify, assess, and mitigate safety and security risks.

\subsection{Example 1: Researcher}

\texttt{Researcher} is a hypothetical agentic AI system which compiles research on a specific topic, similar to OpenAI's or Perplexity's Deep Research. The user provides the research question, then the \texttt{Researcher} clarifies the scope, devises a research plan, searches the web, and compiles the information into a structured report to address the user's question.

Referencing the capabilities in Section \ref{sec:capabilities}, we can identify the \texttt{Researcher}'s capabilities as Planning \& Goal Management, Natural Language Communication, and Internet \& Search Access. Together with the components and design elements and referring to our draft Risk Register in Appendix \ref{app:risk-register-risks}, there are 38 applicable risks to be assessed. We provide an example below:

\fbox{%
  \parbox{0.98\linewidth}{%
    \begin{flushleft}

    \textbf{Risk}: “Opening vulnerabilities to prompt injection attacks via malicious websites” is a \ul{security and safety risk (all)} caused by \ul{external manipulation} of the \ul{Internet \& Search Access capability}.
    \newline
    \newline
    \textbf{Impact}: \textbf{4/5} - Manipulation of the agent can result in a range of safety and security risks that extend beyond the system's boundaries and result in reputational loss for the company.
    \newline
    \textbf{Likelihood}: \textbf{5/5} - Attack has been demonstrated in several real-world case studies, no access to the system required to execute attack.
    \newline
    \newline
    \textbf{Relevance}: \textbf{Relevant} as company's relevance threshold is 3 for impact and 4 for likelihood.
    
    \end{flushleft}
  }%
}

After contextualizing the risks and assessing relevance, only 10 risks remain (RISK-003, RISK-009, RISK-017, RISK-023, RISK-034, RISK-035, RISK-036, RISK-038, RISK-053, and RISK-054) and require technical controls to be implemented for. We provide full explanations in \autoref{app:researcher-example}, rationalizing the impact and likelihood of each risk, and highlighting the relevant risks. Now referring to Appendix \ref{app:risk-register-controls}, there are 17 controls associated with these 10 risks which the team now needs to adopt or adapt to safeguard the agentic system. This step-by-step approach is not only straightforward for developers, but ensures comprehensive understanding of the system's risks.

\subsection{Example 2: Vibe Coder}

\texttt{Vibe Coder} is a hypothetical agentic system which allows non-technical users to develop and deploy simple web apps through natural language prompts, similar to Vercel or Replit. The user specifies the app's key features and design, \texttt{Vibe Coder} proceeds to generate the code and text for the web app, run and create the required front-end and back-end systems locally, and render the website for the user to preview. If the user is satisfied, \texttt{Vibe Coder} will then automatically deploy the web app into a staging environment where it is then ready for user acceptance testing.

Referencing the capabilities in \ref{sec:capabilities}, we can identify quite a few capabilities: Planning \& Goal Management, Tool Use\footnote{Tool use appears only for the \texttt{Vibe Coder} because the agent has the flexibility to choose which tool to accomplish its task, which the research agent does not have (it only has the search tool).}, Natural Language Communication, Internet \& Search Access, Code Execution, File \& Data Management, and System Management. 

Now examining our draft Risk Register in Appendix \ref{app:risk-register-risks}, there are a total of 48 applicable risks - unsurprisingly, this is double the number of capability risks of the \texttt{Researcher}, since there are more capabilities and some of them are also intrinsically riskier. We analyze one risk below:

\fbox{%
  \parbox{0.98\linewidth}{%
    \begin{flushleft}
    \textbf{Risk}: “Overwriting or deleting
database tables or files” is a \ul{security risk (data, application)} caused either by \ul{agent failure or external manipulation} of the \ul{File \& Data Management capability}.
    \newline
    \newline
    \textbf{Impact}: \textbf{3/5} - The app is only deployed into a staging environment and never used in production, but the deletion of files and databases poses a major risk to the system's integrity. 
    \newline
    \textbf{Likelihood}: \textbf{4/5} - Other agentic coding tools like Replit have failed in this manner before \citep{nolan2025replit}, although this is relatively rare and not easily reproduced.
    \newline
    \newline
    \textbf{Relevance}: \textbf{Relevant} as company's relevance threshold is 3 for impact and 3 for likelihood.
    \end{flushleft}
  }%
}

For \texttt{Vibe Coder}, there are a total of 25 relevant risks. This is partly because there are more risks, but also because the company's relevance threshold is lower, arising from a more conservative stance that requires more risks to be directly managed. This results in a much higher number of controls to be included, which is intuitive and sensible given the riskier nature of an agentic coding tool that can execute code and has permissions to modify system resources.

\section{Benefits of the ARC framework}

\textbf{First, the ARC framework enables meaningfully differentiated risk management for different types of agentic systems while still ensuring some level of consistency across all systems.} The component and design elements establish a foundational set of minimum hygiene standards that apply across all agentic systems, guaranteeing a baseline level of safety and security regardless of their specific function or risk profile. Layering on top of that is the capability element, which can vary on the use case and what tools the agent has. This enables a nuanced approach to risk management for agentic systems, as lower-risk systems are not unduly burdened with excessive compliance. 

\textbf{Second, the ARC framework provides forward guidance for developers to build with safety and security considerations upfront, thus avoiding abortive work and encouraging proactivity.} Developers know upfront the risks and controls for each capability, encouraging them to incorporate safety and security considerations into the initial stages of the development lifecycle. By providing clear, actionable guidance upfront, developers can design agentic systems with these safeguards built-in, mitigating risks and reducing developer toil. This also makes the ARC framework more scalable as organizations ramp up adoption of agentic systems across business units and use cases. 

\textbf{Third, the ARC framework has the flexibility to update risks and controls as agentic systems develop and evolve.} The field of agentic AI is characterized by rapid technological advancement and emergent capabilities, leading to an evolving risk landscape. The ARC framework’s systematic risk identification approach helps governance teams make sense of the latest research and real-world incidents and provides a structured way to incorporate the latest risks. The accompanying technical controls can also be refreshed with industry best practices and new tools as they are launched. 

\section{Statement of Contributions}

This paper contributes to ongoing discourse on agentic AI governance with the ARC framework by (1) introducing a novel capability perspective to analyze a wide range of agentic systems; (2) distilling three elements intrinsic to all agentic systems - components, design, and capabilities; (3) establishing a clear nexus between the elements, risks, and controls; and (4) providing a structured and practical approach to help organizations implement the framework. Additionally, we have included a comparison table to other technical governance frameworks for agentic systems in Appendix \ref{app:arc-comparison}.

\section{Conclusion}

As agentic systems become increasingly prevalent, frameworks become essential for safe, ethical, and responsible AI deployment. The ARC framework not only helps organizations manage current risks but also provides a foundation for adapting to future developments in agentic AI capabilities and emerging threat landscapes. With this framework established, future work can focus on developing empirical approaches to validate the risks and controls in the Risk Register and on building automated tools to support the implementation and regular updating of the framework.






\bibliography{iaseai26}


\appendix

\section{Comparison of the ARC framework with other technical governance frameworks}
\label{app:arc-comparison}

In the table below, we compare the ARC framework to three other frameworks (we are regrettably unable to include more due to space limitations).

\newcolumntype{P}[1]{>{\raggedright\arraybackslash}p{#1}}

{
\setlength{\tabcolsep}{4pt}
\renewcommand\arraystretch{1.12}

\rowcolors{2}{gray!10}{white}

\begin{longtable}{P{0.18\textwidth} P{0.19\textwidth} P{0.19\textwidth} P{0.19\textwidth} P{0.19\textwidth}}
\caption{Comparison of ARC framework with alternative frameworks / guides relevant to agentic AI governance and safety.}
\label{tab:arc-compare}\\
\toprule
\textbf{Dimension} &
\textbf{ARC Framework (this paper)} &
\textbf{Dimensional Governance (Engin \& Hand, 2025)} &
\textbf{OWASP Agentic AI (Threats \& Mitigations)} &
\textbf{Google: Secure AI Agents / SAIF 2.0} \\
\midrule
\endfirsthead

\hiderowcolors
\toprule
\textbf{Dimension} &
\textbf{ARC Framework (this paper)} &
\textbf{Dimensional Governance (Engin \& Hand, 2025)} &
\textbf{OWASP Agentic AI (Threats \& Mitigations)} &
\textbf{Google: Secure AI Agents / SAIF 2.0} \\
\midrule
\showrowcolors
\endhead

\hiderowcolors
\midrule
\multicolumn{5}{r}{\small\itshape Continued on next page}\\
\midrule
\showrowcolors
\endfoot

\bottomrule
\endlastfoot


\textbf{Core framing} &
Capability-centric governance mapping \emph{capabilities $\to$ risks $\to$ controls} with structured implementation approach. &
Governance via continuous dimensions and trust thresholds (authority, autonomy, accountability). &
Threat-model of agentic attack surfaces (reasoning, memory, tools, identity, oversight, multi-agent) with mitigations. &
Principles for agents: human controller, limited powers, observable planning / actions; defense-in-depth. \\

\textbf{Primary audience} &
Organizational governance plus product / security teams. &
Policymakers, oversight / governance leads. &
Security engineers, red / blue teams. &
CISOs, security architects, enterprise builders. \\

\textbf{Unit of analysis} &
\textbf{Capabilities} with components and system design. &
\textbf{Dimensions} (continuous scales). &
\textbf{Threats / attack surfaces} for agent workflows. &
\textbf{Principles / control families} across lifecycle. \\

\textbf{Prescriptiveness} &
Medium--High (risk $\to$ control mappings; checklists). &
Low--Medium (conceptual thresholds, fewer concrete controls). &
Medium--High (enumerated threats with mitigations). &
Medium (principle-led control families). \\

\textbf{Coverage of agentic specifics} &
Strong: capability lens tailored to powers / autonomy. &
Conceptual: governance dynamics of agents. &
Strong: agent-specific threats (tools, memory, multi-agent). &
Strong: agent-explicit principles (controller, limits, observability). \\

\textbf{Evidence / evaluation} &
Conceptual plus worked examples; no empirical evaluation yet. &
Conceptual; no empirical evaluation. &
Practitioner-grounded examples; no formal evaluation. &
Policy / engineering narratives; no formal benchmarks. \\

\textbf{Typical artifacts} &
Risk Register; capability profile; control tiers / checklists; sign-off workflow. &
Dimension definitions; threshold guidance; oversight roles. &
Threat navigator; threat / mitigation sheets; red-team prompts. &
Principles plus control families; CISO guidance. \\

\textbf{Control selection logic} &
By \emph{capability profile} and \emph{contextualized relevance threshold} (impact $\times$ likelihood) $\to$ minimum control set. &
By \emph{dimensional thresholds} (e.g., higher autonomy $\Rightarrow$ stricter oversight). &
By \emph{threat presence} (e.g., tool misuse $\Rightarrow$ sandboxing, PoLP). &
By \emph{principles} (limit powers; ensure observability; human controller). \\

\textbf{Verification / testing} &
Adversarial testing; logging / traceability (pre / post metrics recommended). &
Oversight / accountability emphasized; testing not central. &
Threat-led testing / red-teaming against agent surfaces. &
Monitoring / observability of plans / actions emphasized. \\

\textbf{Strengths} &
Holistic, capability-aware; ties risks to controls with governance workflow. &
Clear governance lens; adaptivity via dimensions / thresholds; policy-friendly. &
Security-grounded; actionable mitigations for engineers. &
Enterprise-aligned principles; guardrails for agent power / visibility. \\

\textbf{Gaps} &
Would benefit from empirical evaluation of the approach. &
Less prescriptive; limited implementation detail. &
Governance / process coverage thinner. &
High-level; few concrete test harnesses / metrics. \\

\textbf{Best fit} &
Org-level, cross-functional governance for varied agentic systems. &
Regulators / oversight; policy design and audits. &
Security hardening of agent stacks / tools. &
Enterprise security principles for agent platforms. \\

\end{longtable}
}

\section{Risk Register (Risks)}
\label{app:risk-register-risks}

We provide a preliminary version of a Risk Register below, with a mapping from the element to the risk. Due to space constraints, the controls are presented in a separate table in the next section.

\newcolumntype{M}[1]{>{\centering\arraybackslash}m{#1}}
\newcolumntype{L}[1]{>{\raggedright\arraybackslash}m{#1}}
\renewcommand{\arraystretch}{1.3}
\setlength{\extrarowheight}{2pt}
\renewcommand{\multirowsetup}{\centering\arraybackslash}
\setlength{\LTpre}{0pt}
\setlength{\LTpost}{0pt}
\setlength{\extrarowheight}{1pt}
\begin{longtable}{|p{0.1\textwidth}|p{0.18\textwidth}|p{0.12\textwidth}|p{0.48\textwidth}|}
\rowcolor{black}
\color{white}\textbf{Element} & \color{white}\textbf{Name} & \color{white}\textbf{Risk ID} & \color{white}\textbf{Risk Statement and Description} \\
\hline
\endhead
\multirow{3}{=}{Component}
& \multirow{3}{=}{LLM}
& RISK-001 & \textbf{Use of untrusted or compromised LLMs}: This risk arises when LLMs obtained from untrusted or insufficiently vetted sources have been intentionally poisoned or backdoored during training or distribution, causing them to behave maliciously or unpredictably under specific conditions. Such models may leak sensitive information, bypass safeguards, or execute hidden behaviors that undermine system integrity and trust. \\ \cline{3-4}
& & RISK-002 & \textbf{Insufficient alignment of LLM behaviour}: This risk arises when an LLM's learned objectives and behaviors do not reliably align with intended user goals, system instructions, or organizational policies, leading to inappropriate, unsafe, or undesired outputs. Misalignment may surface as failure to follow constraints, inconsistent reasoning, or behavior that diverges from expected norms in edge cases or complex scenarios. \\ \cline{3-4}
& & RISK-003 & \textbf{Insufficient LLM capability and reliability}: This risk arises when an LLM lacks sufficient capability, robustness, or reasoning performance to correctly interpret instructions, handle edge cases, or detect unsafe situations. As a result, the model may produce incorrect, misleading, or unsafe outputs that create downstream safety or security failures in systems that rely on its judgments. \\ \hline
\multirow{4}{=}{Component}
& \multirow{4}{=}{Tools}
& RISK-004 & \textbf{Weak tool authentication and authorisation controls}: This risk arises when tools connected to an agent lack robust authentication or fine-grained authorisation mechanisms, allowing unauthorised access or misuse of tool capabilities. As a result, attackers or misbehaving agents may compromise the system by invoking sensitive actions, escalating privileges, or manipulating external resources beyond intended boundaries. \\ \cline{3-4}
& & RISK-005 & \textbf{Lack of proper role-based access control for tools}: This risk arises when tools exposed to an agent do not enforce clear, role-based access controls, allowing agents to access capabilities or resources beyond their intended responsibilities. As a result, agents may perform unauthorised actions, misuse sensitive tools, or exceed their permitted scope, increasing the likelihood of security and operational failures. \\ \cline{3-4}
& & RISK-006 & \textbf{Tool poisoning by malicious actors}: This risk arises when tools or their interfaces are intentionally modified, compromised, or replaced by malicious actors to introduce harmful or deceptive behaviour when invoked by an agent. As a result, the agent may unknowingly execute malicious actions, leak sensitive information, or produce manipulated outputs that undermine system integrity and trust. \\ \cline{3-4}
& & RISK-007 & \textbf{Lack of input sanitisation}: This risk arises when inputs passed from the agent to tools are not properly validated or sanitised, allowing malformed or malicious data to be processed. As a result, tools may be exploited through injection attacks, unintended command execution, or data corruption. \\ \hline
\multirow{2}{=}{Component}
& \multirow{2}{=}{Instructions}
& RISK-008 & \textbf{Vague or underspecified instructions}: This risk arises when instructions provided to an LLM are ambiguous, incomplete, or poorly scoped, leading the model to make unintended assumptions when interpreting tasks or constraints. As a result, the LLM may behave unpredictably, bypass safeguards, or take actions that introduce safety or security risks. \\ \cline{3-4}
& & RISK-009 & \textbf{Unsanitised inputs in system instructions}: This risk arises when untrusted or user-controlled inputs are incorporated into system instructions without proper sanitisation or validation. As a result, malicious or malformed content may manipulate the model's behaviour, override intended constraints, or trigger unintended actions. \\ \hline
\multirow{2}{=}{Component}
& \multirow{2}{=}{Memory}
& RISK-010 & \textbf{Poisoned memory}: This risk arises when the memory component of an agentic system is intentionally or inadvertently populated with malicious, misleading, or corrupted information. As a result, the agent may rely on compromised memory to make decisions, propagate false information, or exhibit persistent unsafe behaviour across interactions. \\ \cline{3-4}
& & RISK-011 & \textbf{Sensitive data leakage across memory contexts}: This risk arises when the memory component retains or exposes sensitive information across sessions, tasks, or users with different scopes or authorisations. As a result, data may be inappropriately accessed or reused in unrelated contexts, leading to privacy breaches, confidentiality violations, or unauthorised disclosure. \\ \hline
\multirow{3}{=}{Design}
& \multirow{3}{=}{Agentic Architecture}
& RISK-012 & \textbf{Cascading errors in multi-agent architectures}: This risk arises when errors or misjudgements produced by one agent propagate through interconnected agents within a multi-agent system. As a result, small failures may compound across agent interactions, leading to amplified errors, degraded system performance, or unintended outcomes at the system level. \\ \cline{3-4}
& & RISK-013 & \textbf{Man-in-the-middle attacks between agents}: This risk arises when communication channels between agents are insufficiently secured, allowing an attacker to intercept, modify, or replay messages exchanged within the agentic system. As a result, agents may act on tampered information, leading to incorrect coordination, unauthorised actions, or compromised system behaviour. \\ \cline{3-4}
& & RISK-014 & \textbf{Feedback loops and runaway agent behaviour}: This risk arises when agents repeatedly reinforce each other's decisions, outputs, or errors within an agentic architecture. As a result, feedback loops may form that escalate actions, consume excessive resources, or cause the system to persist in harmful or unintended behaviour without effective human intervention. \\ \hline
\multirow{2}{=}{Design}
& \multirow{2}{=}{Roles and Access Controls}
& RISK-015 & \textbf{Overly permissive roles and permissions}: This risk arises when agents are granted roles or permissions that exceed their intended responsibilities or operational needs. As a result, agents may access sensitive resources, invoke high-impact capabilities, or perform unauthorised actions that increase the likelihood of security, privacy, or operational failures. \\ \cline{3-4}
& & RISK-016 & \textbf{Unauthorised privilege escalation}: This risk arises when agents are able to gain elevated roles or permissions beyond those initially granted, whether through misconfiguration, exploitation, or unintended system behaviour. As a result, agents may bypass intended controls, access restricted resources, or execute actions that undermine system security and governance. \\ \hline
\multirow{2}{=}{Design}
& \multirow{2}{=}{Monitoring and Traceability}
& RISK-017 & \textbf{Delayed failure detection due to limited monitoring}: This risk arises when monitoring systems provide insufficient visibility into agent behaviour, system events, or execution outcomes. As a result, failures, anomalies, or unintended actions may go undetected for extended periods, increasing the impact and difficulty of remediation. \\ \cline{3-4}
& & RISK-018 & \textbf{Inability to audit failures due to missing decision traces}: This risk arises when monitoring systems do not capture sufficient reasoning steps, decision pathways, or execution context for agent actions. As a result, operators may be unable to reconstruct failures, understand why specific outcomes occurred, or conduct effective audits and post-incident reviews. \\ \hline
\multirow{2}{=}{Capability}
& \multirow{2}{=}{Planning and Goal Management}
& RISK-019 & \textbf{Generating plans that fail to meet the user's requirements}: This risk arises when an agent generates plans or goals that do not accurately reflect the user's stated objectives, constraints, or preferences. As a result, the system may pursue incorrect or suboptimal actions, waste resources, or deliver outcomes that do not satisfy user expectations. \\ \cline{3-4}
& & RISK-020 & \textbf{Generating plans that overlook safety implications}: This risk arises when an agent generates plans or goals without adequately considering basic safety, security, or practical constraints that would be apparent to a human. As a result, the system may propose or pursue actions that are unsafe, insecure, or inappropriate despite being technically feasible. \\ \hline
\multirow{2}{=}{Capability}
& \multirow{2}{=}{Agent Delegation}
& RISK-021 & \textbf{Incorrect task delegation between agents}: This risk arises when an agent assigns tasks to other agents that do not match their capabilities, roles, or access permissions. As a result, tasks may be executed incorrectly, fail to complete, or introduce security and operational issues due to inappropriate delegation. \\ \cline{3-4}
& & RISK-022 & \textbf{Malicious or manipulative use of delegated agents}: This risk arises when an agent deliberately assigns tasks to other agents in ways intended to bypass controls, obscure responsibility, or achieve malicious objectives. As a result, delegated agents may be coerced into performing unauthorised actions, amplifying harmful behaviour or evading detection within the system. \\ \hline
\multirow{1}{=}{Capability}
& \multirow{1}{=}{Tool Use}
& RISK-023 & \textbf{Incorrect tool selection or misuse}: This risk arises when an agent selects an inappropriate tool or applies a tool incorrectly for a given task or action. As a result, the agent may produce erroneous outcomes, fail to complete the task effectively, or trigger unintended side effects due to misuse of tool capabilities. \\ \hline
\multirow{6}{=}{Capability}
& \multirow{6}{=}{Multimodal Understanding and Generation}
& RISK-024 & \textbf{Generation of undesirable content}: This risk arises when an agent generates text, images, audio, or other media that contain toxic, hateful, sexual, or otherwise inappropriate content. As a result, the system may cause harm to users, violate organisational standards or regulations, or undermine trust in the system's outputs. \\ \cline{3-4}
& & RISK-025 & \textbf{Generation of unqualified advice in specialised domains}: This risk arises when an agent generates advice or guidance in specialised domains such as medical, financial, or legal contexts without appropriate expertise, validation, or safeguards. As a result, users may act on incorrect or inappropriate information, leading to potential harm or adverse outcomes. \\ \cline{3-4}
& & RISK-026 & \textbf{Generation of controversial or sensitive content}: This risk arises when an agent generates content related to sensitive or controversial topics, such as political commentary or denigrating comments about competitors. As a result, the system may create reputational, legal, or compliance issues, or be perceived as biased, inappropriate, or misrepresentative of organisational views. \\ \cline{3-4}
& & RISK-027 & \textbf{Regurgitating personally identifiable information}: This risk arises when an agent reproduces personally identifiable information in its generated outputs, whether drawn from training data, memory, or prior interactions. As a result, the system may violate privacy obligations, expose individuals to harm, or breach data protection requirements. \\ \cline{3-4}
& & RISK-028 & \textbf{Generation of non-factual or hallucinated content}: This risk arises when an agent generates information that is inaccurate, fabricated, or unsupported by evidence while presenting it as factual. As a result, users may be misled, make incorrect decisions, or lose trust in the system's outputs. \\ \cline{3-4}
& & RISK-029 & \textbf{Generation of copyrighted content}: This risk arises when an agent generates content that reproduces or closely resembles copyrighted material without appropriate rights or attribution. As a result, the system may infringe intellectual property laws, expose the organisation to legal liability, or violate licensing and usage terms. \\ \hline
\multirow{2}{=}{Capability}
& \multirow{2}{=}{Official Communication}
& RISK-030 & \textbf{Misrepresentation of authorship}: This risk arises when recipients are misled about whether an official communication was authored by a human or generated by an agent on behalf of the organisation. As a result, stakeholders may form incorrect assumptions about accountability, intent, or authority, potentially leading to trust, legal, or reputational issues. \\ \cline{3-4}
& & RISK-031 & \textbf{Inaccurate promises or statements in official communications}: This risk arises when an agent makes commitments, assurances, or public statements that are incorrect, unsupported, or exceed the organisation's actual intentions or capabilities. As a result, the organisation may face reputational damage, legal exposure, or loss of public trust due to unmet expectations or misinformation. \\ \hline
\multirow{2}{=}{Capability}
& \multirow{2}{=}{Business Transactions}
& RISK-032 & \textbf{Unauthorised execution of business transactions}: This risk arises when an agent initiates, authorises, or executes business transactions outside predefined approval thresholds, roles, or authorisation limits. As a result, the organisation may be exposed to unintended financial losses, binding contractual obligations, or operational commitments that were not properly sanctioned. \\ \cline{3-4}
& & RISK-033 & \textbf{Leakage of transaction credentials}: This risk arises when credentials, tokens, or sensitive authentication information used to execute business transactions are exposed, mishandled, or improperly stored by an agent or its supporting systems. As a result, malicious parties may gain the ability to initiate unauthorised transactions, manipulate financial operations, or compromise transactional systems. \\ \hline
\multirow{2}{=}{Capability}
& \multirow{2}{=}{Internet and Search Access}
& RISK-034 & \textbf{Prompt injection via malicious websites}: This risk arises when an agent retrieves or processes content from malicious or untrusted websites that are designed to inject instructions or manipulative prompts into the system. As a result, the agent may follow unintended commands, override intended constraints, or take actions that compromise system behaviour or integrity. \\ \cline{3-4}
& & RISK-035 & \textbf{Unreliable information or websites}: This risk arises when an agent retrieves and presents information from websites that are inaccurate, outdated, biased, or otherwise unreliable. As a result, users may be misinformed or make incorrect decisions based on content that has not been adequately validated or corroborated. \\ \hline
\multirow{2}{=}{Capability}
& \multirow{2}{=}{Computer Use}
& RISK-036 & \textbf{Prompt injection risks through computer use}: This risk arises when an agent interacts with graphical user interfaces that display untrusted or adversarial content - such as web pages, documents, pop-ups, or form fields - crafted to embed hidden instructions or manipulative cues. As a result, the agent may misinterpret on-screen text as authoritative guidance, follow injected instructions, or perform unintended actions while operating the interface. \\ \cline{3-4}
& & RISK-037 & \textbf{Exposure of sensitive data}: This risk arises when an agent operating a computer interface accesses websites or applications that contain personally identifiable or sensitive information, particularly when authenticated as a user or organisation. As a result, the agent may inadvertently view, process, or disclose confidential data beyond its intended scope or authorisation. \\ \hline
\multirow{1}{=}{Capability}
& \multirow{1}{=}{Other Programmatic Interfaces}
& RISK-038 & \textbf{Incorrect use of unfamiliar programmatic interfaces}: This risk arises when an agent interacts with programmatic interfaces it has not been trained or configured to use correctly, particularly bespoke or non-standard interfaces outside established protocols such as MCP servers. As a result, the agent may misinterpret interface semantics, invoke operations incorrectly, or produce unintended effects due to improper integration or usage. \\ \hline
\multirow{2}{=}{Capability}
& \multirow{2}{=}{Code Execution}
& RISK-039 & \textbf{Production or execution of poor or ineffective code}: This risk arises when an agent generates or executes code that is incorrect, inefficient, insecure, or unsuitable for the intended task. As a result, the code may fail to achieve desired outcomes, introduce bugs or vulnerabilities, or cause operational disruptions when deployed or run. \\ \cline{3-4}
& & RISK-040 & \textbf{Production or execution of vulnerable or malicious code}: This risk arises when an agent executes code that contains security vulnerabilities or intentionally malicious logic, whether generated by the model or sourced externally. As a result, the system may be compromised through exploitation, unauthorised access, data leakage, or other harmful effects. \\ \hline
\pagebreak
\multirow{4}{=}{Capability}
& \multirow{4}{=}{File and Data Management}
& RISK-041 & \textbf{Unintended overwriting or deletion of files or data}: This risk arises when an agent modifies, overwrites, or deletes files, database tables, or datasets without explicit user instruction or authorisation. As a result, critical information may be lost or corrupted, leading to data integrity issues, operational disruption, or the need for costly recovery efforts. \\ \cline{3-4}
& & RISK-042 & \textbf{Database overload due to inefficient data operations}: This risk arises when an agent issues poorly optimised, excessively frequent, or redundant queries against databases or data stores. As a result, system performance may degrade, resources may be exhausted, or critical services may become unavailable due to unnecessary load. \\ \cline{3-4}
& & RISK-043 & \textbf{Exposure of sensitive data through file or database access}: This risk arises when an agent accesses, processes, or outputs personally identifiable or sensitive information stored in files or databases without appropriate safeguards. As a result, confidential data may be disclosed to unauthorised parties, leading to privacy breaches, regulatory non-compliance, or loss of trust. \\ \cline{3-4}
& & RISK-044 & \textbf{Prompt injection via malicious files or data}: This risk arises when an agent ingests or processes maliciously crafted files or data that embed hidden instructions or manipulative content. As a result, the agent may follow unintended prompts, alter its behaviour, or execute actions that compromise system safety or integrity. \\ \hline
\multirow{2}{=}{Capability}
& \multirow{2}{=}{System Management}
& RISK-045 & \textbf{Misconfiguration of system resources}: This risk arises when an agent incorrectly configures system settings, infrastructure resources, or operational parameters. As a result, system performance, reliability, or security may be degraded, leading to service disruptions or unintended operational behaviour. \\ \cline{3-4}
& & RISK-046 & \textbf{System overload due to inefficient or excessive operations}: This risk arises when an agent issues poorly optimised, excessively frequent, or redundant system-level operations or queries. As a result, computing resources may be exhausted, system performance may degrade, or services may become unavailable due to unnecessary load. \\ \hline
\end{longtable}

\section{Risk Register (Controls)}
\label{app:risk-register-controls}

We provide a preliminary version of a Risk Register below, with a mapping from each risk to a control. Due to space constraints, the elements to risk mappings are presented in a separate table in the previous section. Control levels indicate criticality: Level 0 (Cardinal) are fundamental requirements, Level 1 (Standard) should be adopted or adapted meaningfully, and Level 2 (Best Practice) are recommended for high-risk systems.

\begin{longtable}{|M{0.10\textwidth}|M{0.26\textwidth}|M{0.10\textwidth}|M{0.06\textwidth}|M{0.36\textwidth}|}
\rowcolor{black}
\color{white}\textbf{Risk ID} & \color{white}\textbf{Risk Statement} & \color{white}\textbf{Control ID} & \color{white}\textbf{Level} & \color{white}\textbf{Control Statement} \\
\hline
\endfirsthead

\rowcolor{black}
\color{white}\textbf{Risk ID} & \color{white}\textbf{Risk Statement} & \color{white}\textbf{Control ID} & \color{white}\textbf{Level} & \color{white}\textbf{Control Statement} \\
\hline
\endhead
\multirow{5}{=}{RISK-001}
& \multirow{5}{=}{Use of untrusted or compromised LLMs}
& CTRL-0001 & 0 & Use only LLMs from verified and trusted model developers \\ \cline{3-5}
& & CTRL-0002 & 0 & Obtain legally binding no-training and no-logging agreements from LLM API service providers \\ \cline{3-5}
& & CTRL-0003 & 1 & Use only established and verified model loaders in production environments \\ \cline{3-5}
& & CTRL-0006 & 1 & Require human approval before executing high-impact actions \\ \cline{3-5}
& & CTRL-0007 & 0 & Log all LLM inputs and outputs for regular review \\ \hline
\multirow{5}{=}{RISK-002}
& \multirow{5}{=}{Insufficient alignment of LLM behaviour}
& CTRL-0004 & 2 & Review the LLM's system card to inform risk assessment and model selection \\ \cline{3-5}
& & CTRL-0005 & 0 & Conduct structured evaluation of multiple LLMs for instruction-following, performance, and safety before deployment \\ \cline{3-5}
& & CTRL-0006 & 1 & Require human approval before executing high-impact actions \\ \cline{3-5}
& & CTRL-0007 & 0 & Log all LLM inputs and outputs for regular review \\ \cline{3-5}
& & CTRL-0008 & 1 & Implement automated alerts when agent behaviour drifts from predefined thresholds \\ \hline
\multirow{4}{=}{RISK-003}
& \multirow{4}{=}{Insufficient LLM capability and reliability}
& CTRL-0004 & 2 & Review the LLM's system card to inform risk assessment and model selection \\ \cline{3-5}
& & CTRL-0005 & 0 & Conduct structured evaluation of multiple LLMs for instruction-following, performance, and safety before deployment \\ \cline{3-5}
& & CTRL-0006 & 1 & Require human approval before executing high-impact actions \\ \cline{3-5}
& & CTRL-0007 & 0 & Log all LLM inputs and outputs for regular review \\ \hline
\multirow{3}{=}{RISK-004}
& \multirow{3}{=}{Weak tool authentication and authorisation controls}
& CTRL-0009 & 0 & Use only MCP servers that implement robust authentication mechanisms in production environments \\ \cline{3-5}
& & CTRL-0010 & 1 & Use only MCP servers that validate credentials on every inbound request \\ \cline{3-5}
& & CTRL-0032 & 0 & Centralise observability data collection in a unified backend system \\ \hline
\multirow{2}{=}{RISK-005}
& \multirow{2}{=}{Lack of proper role-based access control for tools}
& CTRL-0011 & 0 & Limit token scopes to the minimum privileges required and avoid broad or wildcard scopes \\ \cline{3-5}
& & CTRL-0012 & 2 & Use only MCP servers that integrate with authorisation servers implementing per-client consent mechanisms \\ \hline
\multirow{2}{=}{RISK-006}
& \multirow{2}{=}{Tool poisoning by malicious actors}
& CTRL-0013 & 0 & Test all untested MCP servers in a sandboxed environment before deploying to production \\ \cline{3-5}
& & CTRL-0014 & 0 & Use only MCP servers from verified and trusted developers \\ \hline
RISK-007
& Lack of input sanitisation
& CTRL-0015 & 1 & Treat all tool metadata and outputs as untrusted input requiring validation \\ \hline
\multirow{3}{=}{RISK-008}
& \multirow{3}{=}{Vague or underspecified instructions}
& CTRL-0016 & 0 & Define clearly the agent's role, scope, and non-goals in the system prompt \\ \cline{3-5}
& & CTRL-0017 & 1 & Define clear success criteria for the agent's tasks \\ \cline{3-5}
& & CTRL-0018 & 2 & Define default behaviour when the agent encounters ambiguous situations \\ \hline
\multirow{2}{=}{RISK-009}
& \multirow{2}{=}{Unsanitised inputs in system instructions}
& CTRL-0019 & 0 & Use delimiters to enclose untrusted inputs and instruct the LLM to treat delimited content as data only \\ \cline{3-5}
& & CTRL-0020 & 2 & Use a dedicated LLM to extract required fields from inputs and filter out extraneous text or embedded instructions \\ \hline
\multirow{3}{=}{RISK-010}
& \multirow{3}{=}{Poisoned memory}
& CTRL-0021 & 0 & Implement allowlists and denylists to restrict what categories of information can be written to agent memory \\ \cline{3-5}
& & CTRL-0022 & 1 & Implement content filtering on memory writes to detect and block known unsafe content patterns \\ \cline{3-5}
& & CTRL-0023 & 2 & Log all memory modifications with comprehensive source metadata for audit purposes \\ \hline
\multirow{2}{=}{RISK-011}
& \multirow{2}{=}{Sensitive data leakage across memory contexts}
& CTRL-0021 & 0 & Implement allowlists and denylists to restrict what categories of information can be written to agent memory \\ \cline{3-5}
& & CTRL-0023 & 2 & Log all memory modifications with comprehensive source metadata for audit purposes \\ \hline
\multirow{2}{=}{RISK-012}
& \multirow{2}{=}{Cascading errors in multi-agent architectures}
& CTRL-0024 & 0 & Define formal schemas for inter-agent messages and validate all messages against these schemas before processing \\ \cline{3-5}
& & CTRL-0025 & 1 & Ensure all inter-agent communications are encrypted in transit and prohibit plaintext channels \\ \hline
\multirow{2}{=}{RISK-013}
& \multirow{2}{=}{Man-in-the-middle attacks between agents}
& CTRL-0026 & 1 & Require all agents to authenticate with verifiable, cryptographically signed identities before processing requests \\ \cline{3-5}
& & CTRL-0027 & 1 & Implement circuit breakers to prevent cascading failures in multi-agent systems \\ \hline
\multirow{2}{=}{RISK-014}
& \multirow{2}{=}{Feedback loops and runaway agent behaviour}
& CTRL-0028 & 0 & Continuously monitor multi-agent systems for cascade failure indicators \\ \cline{3-5}
& & CTRL-0029 & 1 & Grant agents only the minimum permissions required for their designated tasks \\ \hline
\multirow{2}{=}{RISK-015}
& \multirow{2}{=}{Overly permissive roles and permissions}
& CTRL-0030 & 1 & Assign each agent a unique, verifiable identity with no shared credentials \\ \cline{3-5}
& & CTRL-0031 & 1 & Use only MCP servers that validate token provenance and prohibit unauthorised token passthrough \\ \hline
\multirow{2}{=}{RISK-016}
& \multirow{2}{=}{Unauthorised privilege escalation}
& CTRL-0030 & 1 & Assign each agent a unique, verifiable identity with no shared credentials \\ \cline{3-5}
& & CTRL-0032 & 0 & Centralise observability data collection in a unified backend system \\ \hline
\multirow{2}{=}{RISK-017}
& \multirow{2}{=}{Delayed failure detection due to limited monitoring}
& CTRL-0033 & 0 & Standardise trace attributes for agent operations using consistent semantic conventions \\ \cline{3-5}
& & CTRL-0035 & 2 & Require agents to decompose user goals into explicit sub-goals and validate necessity before proceeding \\ \hline
\multirow{2}{=}{RISK-018}
& \multirow{2}{=}{Inability to audit failures due to missing decision traces}
& CTRL-0034 & 0 & Conduct regular reviews of logs and traces to detect emergent issues in deployed agentic systems \\ \cline{3-5}
& & CTRL-0035 & 2 & Require agents to decompose user goals into explicit sub-goals and validate necessity before proceeding \\ \hline
\multirow{3}{=}{RISK-019}
& \multirow{3}{=}{Generating plans that fail to meet the user's requirements}
& CTRL-0006 & 1 & Require human approval before executing high-impact actions \\ \cline{3-5}
& & CTRL-0036 & 1 & Regularly evaluate and test planning behaviour under representative workloads and failure scenarios \\ \cline{3-5}
& & CTRL-0037 & 1 & Require planning agents to include explicit safety constraints in all generated plans before execution \\ \hline
\multirow{3}{=}{RISK-020}
& \multirow{3}{=}{Generating plans that overlook safety implications}
& CTRL-0006 & 1 & Require human approval before executing high-impact actions \\ \cline{3-5}
& & CTRL-0038 & 0 & Conduct pre-deployment safety verification using domain-relevant stress tests and adversarial scenarios \\ \cline{3-5}
& & CTRL-0039 & 1 & Ensure each agent publishes standardised, machine-readable capability descriptors accessible to other agents \\ \hline
RISK-021
& Incorrect task delegation between agents
& CTRL-0040 & 0 & Limit the scope of agent actions through predefined thresholds and baselines \\ \hline
\multirow{4}{=}{RISK-022}
& \multirow{4}{=}{Malicious or manipulative use of delegated agents}
& CTRL-0008 & 1 & Implement automated alerts when agent behaviour drifts from predefined thresholds \\ \cline{3-5}
& & CTRL-0024 & 0 & Define formal schemas for inter-agent messages and validate all messages against these schemas before processing \\ \cline{3-5}
& & CTRL-0025 & 1 & Ensure all inter-agent communications are encrypted in transit and prohibit plaintext channels \\ \cline{3-5}
& & CTRL-0041 & 0 & Provide comprehensive descriptions for each tool including intended use, required inputs, and potential outputs \\ \hline
\multirow{3}{=}{RISK-023}
& \multirow{3}{=}{Incorrect tool selection or misuse}
& CTRL-0042 & 0 & Require explicit human confirmation before executing high-impact or irreversible tool actions \\ \cline{3-5}
& & CTRL-0043 & 1 & Log all tool selection decisions and invocations with comprehensive metadata \\ \cline{3-5}
& & CTRL-0044 & 1 & Implement output safety guardrails to detect and prevent generation of undesirable content \\ \hline
RISK-024
& Generation of undesirable content
& CTRL-0045 & 0 & Implement input guardrails to detect and decline requests for specialised domain advice \\ \hline
RISK-025
& Generation of unqualified advice in specialised domains
& CTRL-0046 & 0 & Implement input guardrails to detect and decline requests for controversial content that violates organisational policies \\ \hline
RISK-026
& Generation of controversial or sensitive content
& CTRL-0047 & 0 & Implement output guardrails to detect and redact personally identifiable information \\ \hline
RISK-027
& Regurgitating personally identifiable information
& CTRL-0048 & 2 & Implement methods to reduce hallucination rates in agent outputs \\ \hline
\multirow{3}{=}{RISK-028}
& \multirow{3}{=}{Generation of non-factual or hallucinated content}
& CTRL-0049 & 0 & Implement UI/UX cues to communicate the risk of hallucination to users \\ \cline{3-5}
& & CTRL-0050 & 1 & Implement features enabling users to verify generated answers against source content \\ \cline{3-5}
& & CTRL-0051 & 0 & Implement input guardrails to detect and decline requests to generate copyrighted content \\ \hline
RISK-029
& Generation of copyrighted content
& CTRL-0052 & 2 & Declare upfront that communications are generated by an AI system \\ \hline
RISK-030
& Misrepresentation of authorship
& CTRL-0053 & 0 & Require human approval for communications on sensitive matters \\ \hline
\multirow{3}{=}{RISK-031}
& \multirow{3}{=}{Inaccurate promises or statements in official communications}
& CTRL-0054 & 0 & Limit agent communications to standard processes with predefined templates \\ \cline{3-5}
& & CTRL-0055 & 1 & Provide alternative channels for users to clarify communications or provide feedback \\ \cline{3-5}
& & CTRL-0056 & 1 & Require explicit user confirmation before initiating or committing any business transaction \\ \hline
\multirow{2}{=}{RISK-032}
& \multirow{2}{=}{Unauthorised execution of business transactions}
& CTRL-0057 & 2 & Require out-of-band confirmation when transaction risk signals are elevated \\ \cline{3-5}
& & CTRL-0058 & 1 & Restrict agents to proposing transactions whilst using a separate transaction controller for execution \\ \hline
\multirow{2}{=}{RISK-033}
& \multirow{2}{=}{Leakage of transaction credentials}
& CTRL-0059 & 2 & Apply fraud detection models or heuristics to agent-proposed transactions \\ \cline{3-5}
& & CTRL-0060 & 1 & Implement escape filtering before incorporating web content into prompts \\ \hline
\multirow{3}{=}{RISK-034}
& \multirow{3}{=}{Prompt injection via malicious websites}
& CTRL-0061 & 0 & Use structured retrieval APIs for web searches rather than web scraping \\ \cline{3-5}
& & CTRL-0062 & 0 & Implement input guardrails to detect prompt injection and adversarial attacks \\ \cline{3-5}
& & CTRL-0063 & 1 & Prioritise search results from verified, high-quality domains \\ \hline
RISK-035
& Unreliable information or websites
& CTRL-0064 & 1 & Limit computer use to accessing only safe and trusted resources \\ \hline
\multirow{2}{=}{RISK-036}
& \multirow{2}{=}{Prompt injection risks through computer use}
& CTRL-0065 & 0 & Ensure computer use capabilities provide immediate interruptability \\ \cline{3-5}
& & CTRL-0066 & 0 & Ensure "take over" mode is activated when entering sensitive data \\ \hline
RISK-037
& Exposure of sensitive data
& CTRL-0067 & 0 & Ensure proper documentation of programmatic interfaces for agent use \\ \hline
RISK-038
& Incorrect use of unfamiliar programmatic interfaces
& CTRL-0068 & 0 & Use code linters to screen generated code for bad practices and poor syntax \\ \hline
\multirow{5}{=}{RISK-039}
& \multirow{5}{=}{Production or execution of poor or ineffective code}
& CTRL-0069 & 0 & Run agent-generated code only in isolated compute environments with network access blocked by default \\ \cline{3-5}
& & CTRL-0070 & 0 & Review all agent-generated code before execution \\ \cline{3-5}
& & CTRL-0071 & 0 & Use static code analysers to detect security vulnerabilities and code quality issues \\ \cline{3-5}
& & CTRL-0072 & 1 & Monitor runtime and memory consumption of agent-generated code \\ \cline{3-5}
& & CTRL-0073 & 0 & Create a denylist of commands that agents are not permitted to execute \\ \hline
\multirow{6}{=}{RISK-040}
& \multirow{6}{=}{Production or execution of vulnerable or malicious code}
& CTRL-0070 & 0 & Review all agent-generated code before execution \\ \cline{3-5}
& & CTRL-0071 & 0 & Use static code analysers to detect security vulnerabilities and code quality issues \\ \cline{3-5}
& & CTRL-0072 & 1 & Monitor runtime and memory consumption of agent-generated code \\ \cline{3-5}
& & CTRL-0074 & 0 & Conduct CVE scanning and block execution of code with High or Critical vulnerabilities \\ \cline{3-5}
& & CTRL-0075 & 1 & Do not grant write access to agents unless strictly necessary \\ \cline{3-5}
& & CTRL-0076 & 1 & Require human approval for any destructive changes to databases, tables, or files \\ \hline
\multirow{3}{=}{RISK-041}
& \multirow{3}{=}{Unintended overwriting or deletion of files or data}
& CTRL-0077 & 0 & Enable versioning or soft-delete for managed object stores to allow recovery from accidental modifications \\ \cline{3-5}
& & CTRL-0078 & 0 & Enforce throttling or rate limits on agent-initiated database operations \\ \cline{3-5}
& & CTRL-0079 & 2 & Validate agent-generated database queries for efficiency before execution against production databases \\ \hline
\multirow{3}{=}{RISK-042}
& \multirow{3}{=}{Database overload due to inefficient data operations}
& CTRL-0080 & 0 & Implement caching mechanisms to reduce repetitive database queries by agents \\ \cline{3-5}
& & CTRL-0081 & 1 & Implement input guardrails to detect personally identifiable information in data accessed by agents \\ \cline{3-5}
& & CTRL-0082 & 2 & Do not grant agents access to personally identifiable or sensitive data unless strictly required \\ \hline
\multirow{2}{=}{RISK-043}
& \multirow{2}{=}{Exposure of sensitive data through file or database access}
& CTRL-0083 & 0 & Disallow unknown or external files unless they have been scanned for threats \\ \cline{3-5}
& & CTRL-0084 & 0 & Set minimum and maximum limits on what agents can modify within system resources \\ \hline
\multirow{2}{=}{RISK-044}
& \multirow{2}{=}{Prompt injection via malicious files or data}
& CTRL-0063 & 1 & Prioritise search results from verified, high-quality domains \\ \cline{3-5}
& & CTRL-0085 & 0 & Log system health metrics and implement automated alerts for abnormal conditions \\ \hline
\multirow{2}{=}{RISK-045}
& \multirow{2}{=}{Misconfiguration of system resources}
& CTRL-0086 & 0 & Limit the number of concurrent queries to external systems by agents \\ \cline{3-5}
& & CTRL-0087 & 0 & Ensure logging of system health metrics and automated alerts to the developer team if any metrics are abnormal \\ \hline
RISK-046
& System overload due to inefficient or excessive operations
& CTRL-0088 & 0 & Limit the number of concurrent queries to external systems from the agent \\ \hline
\end{longtable}

\section{Worked Example: Researcher}
\label{app:researcher-example}

To demonstrate how contextualization works, we fill out the assessment in the Risk Register for the \texttt{Researcher} example here. Note that the applicable relevance threshold is 3 for impact and 4 for likelihood. We highlight in {\color{red}red} risks that exceed the relevance threshold, requiring additional controls to be implemented. 
\newline

\begin{longtable}{|M{0.15\textwidth}|M{0.10\textwidth}|M{0.32\textwidth}|L{0.32\textwidth}|}
\rowcolor{black}
\color{white}\textbf{Category} & \color{white}\textbf{Risk ID} & \color{white}\textbf{Risk Description} & \color{white}\textbf{Assessment} \\
\hline
\endfirsthead

\rowcolor{black}
\color{white}\textbf{Category} & \color{white}\textbf{Risk ID} & \color{white}\textbf{Risk Description} & \color{white}\textbf{Assessment} \\
\hline
\endhead
\multirow{3}{=}{LLM}
& RISK-001 & Use of untrusted or compromised LLMs & \textbf{Impact: 3/5} - Using compromised LLMs could lead to data leakage or manipulation of research outputs.\newline \textbf{Likelihood: 1/5} - Current implementation uses verified LLMs from trusted providers.\newline \newline \textbf{Relevance}: \textbf{Not Relevant}\\ \cline{2-4}
& RISK-002 & Insufficient alignment of LLM behaviour & \textbf{Impact: 2/5} - Researcher is relatively low-stakes as its outputs are viewed and verified by a human-in-the-loop. \newline \textbf{Likelihood: 1/5} - Task is quite narrowly scoped such that it is unlikely to perform non-research tasks. \newline \newline \textbf{Relevance}: \textbf{Not Relevant}\\ \cline{2-4}
& \color{red}RISK-003 & \color{red}Insufficient LLM capability and reliability & \color{red}
    \begin{flushleft}
    \textbf{Impact}: \textbf{4/5} - Insufficient capability means higher likelihood of poor reasoning, incorrect outputs, and safety hazards.
    \newline
    \textbf{Likelihood}: \textbf{4/5} - Research has demonstrated that weaker LLMs are more prone to errors and prompt injection attacks, especially when handling long context which is common in Researcher.
    \newline
    \newline
    \textbf{Relevance}: \textbf{Relevant}
    \end{flushleft}
    \\ \hline
\multirow{2}{=}{Instructions}
& RISK-008 & Vague or underspecified instructions & \textbf{Impact: 1/5} - Users can re-run Researcher requests with step-by-step instructions if wrong actions are taken. \newline \textbf{Likelihood: 1/5} - Researcher clarifies broad steps it will be taking (i.e., research directions) before proceeding.\newline \newline \textbf{Relevance}: \textbf{Not Relevant}\\ \cline{2-4}
& RISK-009 & Unsanitised inputs in system instructions & \textbf{Impact: 2/5} - While Researcher may be used for non-intended purposes, user outputs are consumed by the user only, limiting the impact.  \newline \textbf{Likelihood: 1/5} - Unlikely since delimiters are used to segregate system and user prompts.\newline \newline \textbf{Relevance}: \textbf{Not Relevant}\\ \hline
\multirow{4}{=}{Tools}
& RISK-004 & Weak tool authentication and authorisation controls &
    \begin{flushleft}
    \textbf{Impact}: \textbf{1/5} - No privileged actions for this agentic system
    \newline
    \textbf{Likelihood}: \textbf{1/5} - Current implementation relies on trustworthy Internet search tools like DuckDuckGo.
    \newline
    \newline
    \textbf{Relevance}: \textbf{Not relevant}
    \end{flushleft}
    \\ \cline{2-4}
& RISK-005 & Lack of proper role-based access control for tools &
    \textbf{Impact: 2/5} - No sensitive data stored in the agent, but such tools may allow for lateral access.
    \newline \textbf{Likelihood: 2/5} - Current implementation relies on trustworthy Internet search tools like DuckDuckGo.\newline \newline \textbf{Relevance}: \textbf{Not Relevant}
\\ \cline{2-4}
& RISK-006 & Tool poisoning by malicious actors & \textbf{Impact: 2/5} - Malicious code could shut down the process but each request is processed in an isolated container, reducing its impact on host system. \newline \textbf{Likelihood: 1/5} - Current implementation relies on trustworthy Internet search tools like DuckDuckGo.\newline \newline \textbf{Relevance}: \textbf{Not Relevant}\\ \cline{2-4}
& \color{red}RISK-007 & \color{red}Lack of input sanitisation &  \color{red}
    \begin{flushleft}
    \textbf{Impact}: \textbf{4/5} - Lack of input sanitation means higher likelihood of a jailbreak being passed to the LLM, leading to safety and security hazards.
    \newline
    \textbf{Likelihood}: \textbf{5/5} - Tools without input sanitation have been demonstrated to be particularly susceptible to even simple prompt injection attacks.
    \newline
    \newline
    \textbf{Relevance}: \textbf{Relevant}
    \end{flushleft}
\\ \hline
\multirow{2}{=}{Memory}
& RISK-010 & Poisoned memory & \textbf{Impact: 3/5} - Incorrect data or facts can lead to inaccurate research, but users are expected to check outputs.\newline \textbf{Likelihood: 1/5} - Memory store is protected and can only be updated by authorised system owner, and Researcher tasks tend to be single-turn completions. \newline \newline \textbf{Relevance}: \textbf{Not Relevant}\\ \cline{2-4}
& RISK-011 & Sensitive data leakage across memory contexts & \textbf{Impact: 1/5} - Prior interactions are stored in a separate database, and not provided to the agent at runtime \newline \textbf{Likelihood: 1/5} - Researcher tasks tend to be open-ended queries and do not include specific information. \newline \newline \textbf{Relevance}: \textbf{Not Relevant}\\ \hline
\multirow{3}{=}{Agentic Architecture}
& \color{red}RISK-012 & \color{red}Cascading errors in multi-agent architectures & \color{red}
    \begin{flushleft}
    \textbf{Impact}: \textbf{4/5} - Success of the research depends a lot on the research questions and direction being set by the first agent, and the agent which summarizes the information depends on accurate data being returned by the web search agent.
    \newline
    \textbf{Likelihood}: \textbf{4/5} - Architecture does not require reflexive checking of statements from prior agents, making this quite a likely risk.
    \newline
    \newline
    \textbf{Relevance}: \textbf{Relevant}
    \end{flushleft}
    \\ \cline{2-4}
& RISK-013 & Man-in-the-middle attacks between agents &
    \begin{flushleft}
    \textbf{Impact}: \textbf{2/5} - Even if such attacks occur, only system integrity will be affected, but there are no sensitive data in the agentic system.
    \newline
    \textbf{Likelihood}: \textbf{1/5} - Unlikely that MitM attacks will succeed due to the security of the A2A protocol
    \newline
    \newline
    \textbf{Relevance}: \textbf{Not relevant}
    \end{flushleft}
    \\ \cline{2-4}
& RISK-018 & Inability to audit failures due to missing decision traces &
    \begin{flushleft}
    \textbf{Impact}: \textbf{1/5} - Logging is straightforward for this system
    \newline
    \textbf{Likelihood}: \textbf{2/5} - Linear architecture makes reconstruction of past reasoning traces relatively easy
    \newline
    \newline
    \textbf{Relevance}: \textbf{Not relevant}
    \end{flushleft}
    \\ \hline
\multirow{2}{=}{Roles and Access Controls}
& RISK-015 & Overly permissive roles and permissions &
    \begin{flushleft}
    \textbf{Impact}: \textbf{1/5} - No restricted resources, so overly permissive roles are unlikely to have much effect.
    \newline
    \textbf{Likelihood}: \textbf{1/5} - No access controls granted over internal restricted data or files
    \newline
    \newline
    \textbf{Relevance}: \textbf{Not relevant}
    \end{flushleft}
    \\ \cline{2-4}
& RISK-016 & Unauthorised privilege escalation &
    \begin{flushleft}
    \textbf{Impact}: \textbf{1/5} - No restricted resources
    \newline
    \textbf{Likelihood}: \textbf{1/5} - No access controls granted over internal restricted data or files
    \newline
    \newline
    \textbf{Relevance}: \textbf{Not relevant}
    \end{flushleft}
    \\ \hline
\multirow{2}{=}{Monitoring and Traceability}
& RISK-017 & Delayed failure detection due to limited monitoring &
    \begin{flushleft}
    \textbf{Impact}: \textbf{2/5} - Impact is largely limited to the system, and there are user feedback channels available
    \newline
    \textbf{Likelihood}: \textbf{2/5} - Generally rare unless there are wider outages
    \newline
    \newline
    \textbf{Relevance}: \textbf{Not relevant}
    \end{flushleft}
    \\ \cline{2-4}
& RISK-018 & Inability to audit failures due to missing decision traces &
    \begin{flushleft}
    \textbf{Impact}: \textbf{2/5} - Largely confined to the system-level, and decision-making for this system is linear so it is easier to trace.
    \newline
    \textbf{Likelihood}: \textbf{2/5} - Possible in theory, but no demonstrated failures so far for the research agent
    \newline
    \newline
    \textbf{Relevance}: \textbf{Not relevant}
    \end{flushleft}
    \\ \hline
\multirow{2}{=}{Planning and Goal Management}
& RISK-019 & Generating plans that fail to meet the user's requirements &
    \begin{flushleft}
    \textbf{Impact}: \textbf{3/5} - Largely confined to the system-level as users will simply stop using the agent if it produces poor-quality outputs
    \newline
    \textbf{Likelihood}: \textbf{2/5} - Generally rare unless the topic is highly specialized
    \newline
    \newline
    \textbf{Relevance}: \textbf{Not relevant}
    \end{flushleft}
    \\ \cline{2-4}
& RISK-020 & Generating plans that overlook safety implications &
    \begin{flushleft}
    \textbf{Impact}: \textbf{1/5} - Largely confined to the system-level as users will simply stop using the agent if it produces nonsensical outputs
    \newline
    \textbf{Likelihood}: \textbf{2/5} - Generally rare unless the topic is highly specialized
    \newline
    \newline
    \textbf{Relevance}: \textbf{Not relevant}
    \end{flushleft}
    \\ \hline
\multirow{6}{=}{Multimodal Understanding and Generation}
& \color{red}RISK-024 & \color{red}Generation of undesirable content &     \color{red}
    \begin{flushleft}
    \textbf{Impact}: \textbf{4/5} - Undesirable content may shock and offend users, especially for particularly discriminatory or NSFW content.
    \newline
    \textbf{Likelihood}: \textbf{4/5} - LLMs and agents have been demonstrated to be susceptible to attacks to generate such undesirable content.
    \newline
    \newline
    \textbf{Relevance}: \textbf{Relevant}
    \end{flushleft}
    \\ \cline{2-4}
& \color{red}RISK-025 & \color{red}Generation of unqualified advice in specialised domains & \color{red}
    \begin{flushleft}
    \textbf{Impact}: \textbf{4/5} - As the \texttt{Researcher} is meant to help people do deep research into specific, complex topics, users are likely to trust the outputs even if the advice is unqualified, which in turn may result in significant safety and liability concerns.
    \newline
    \textbf{Likelihood}: \textbf{4/5} - While most LLMs tend to qualify their statements and ask users to seek professional advice, they still provide their advice to the users anyway.
    \newline
    \newline
    \textbf{Relevance}: \textbf{Relevant}
    \end{flushleft}
    \\ \cline{2-4}
& \color{red}RISK-026 & \color{red}Generation of controversial or sensitive content & \color{red}
    \begin{flushleft}
    \textbf{Impact}: \textbf{3/5} - As the system will take in any research topic by the user, it is plausible for a user to ask the research agent to do research into controversial topics.
    \newline
    \textbf{Likelihood}: \textbf{4/5} - While this has not been systematically demonstrated in academic papers, it is relatively easy to carry this out even in standard research agent applications.
    \newline
    \newline
    \textbf{Relevance}: \textbf{Relevant}
    \end{flushleft}
    \\ \cline{2-4}
& RISK-027 & Regurgitating personally identifiable information &
    \begin{flushleft}
    \textbf{Impact}: \textbf{2/5} - As the agentic system will only search the internet, any information returned to the user has to be discoverable on the public internet, which limits the liability and impact of PII regurgitation.
    \newline
    \textbf{Likelihood}: \textbf{3/5} - While there have been several papers demonstrating attacks on LLMs for regurgitation of PII, they have not been proven to succeed against agentic systems and web search agents.
    \newline
    \newline
    \textbf{Relevance}: \textbf{Not relevant}
    \end{flushleft}
    \\ \cline{2-4}
& \color{red}RISK-028 & \color{red}Generation of non-factual or hallucinated content & \color{red}
    \begin{flushleft}
    \textbf{Impact}: \textbf{4/5} - Providing factually accurate information is a core feature of this agentic system.
    \newline
    \textbf{Likelihood}: \textbf{5/5} - Several studies have shown that LLMs have a strong tendency to hallucinate, especially for highly specialized topics which users are likely to ask research agentic systems about.
    \newline
    \newline
    \textbf{Relevance}: \textbf{Relevant}
    \end{flushleft}
    \\ \cline{2-4}
& RISK-029 & Generation of copyrighted content &
    \begin{flushleft}
    \textbf{Impact}: \textbf{3/5} - Potential legal liability if copyrighted content is reproduced by the system
    \newline
    \textbf{Likelihood}: \textbf{2/5} - While LLMs have been shown to reproduce copyrighted content in full when asked for, this has not been demonstrated for web search agents which have to summarize multiple pages.
    \newline
    \newline
    \textbf{Relevance}: \textbf{Not relevant}
    \end{flushleft}
    \\ \hline
\multirow{2}{=}{Internet and Search Access}
& \color{red}RISK-034 & \color{red}Prompt injection via malicious websites & \color{red}
    \begin{flushleft}
    \textbf{Impact}: \textbf{4/5} - Manipulation of the agent can result in a range of safety and security risks that extend beyond the system's boundaries and result in reputational loss for the company.
    \newline
    \textbf{Likelihood}: \textbf{5/5} - Attack has been demonstrated in several real-world case studies, no access to the system required to execute attack.
    \newline
    \newline
    \textbf{Relevance}: \textbf{Relevant}
    \end{flushleft}
    \\ \cline{2-4}
& \color{red}RISK-035 & \color{red}Unreliable information or websites & \color{red}
    \begin{flushleft}
    \textbf{Impact}: \textbf{4/5} - Summarizing reliable and accurate information is a core feature for the system and cannot be compromised on.
    \newline
    \textbf{Likelihood}: \textbf{5/5} - Several real-world examples have demonstrated LLMs sometimes return statements from satirical websites as the truth.
    \newline
    \newline
    \textbf{Relevance}: \textbf{Relevant}
    \end{flushleft}
    \\ \hline
\end{longtable}

\end{document}